\pdfoutput=1

\documentclass[11pt]{article}

\newif\ifarxiv
\arxivtrue

\ifarxiv
    \usepackage{acl}
\else
    \usepackage[review]{acl}
\fi

\usepackage{times}
\usepackage{latexsym}
\usepackage{amsmath,amssymb,amsthm}
\usepackage{bbm}
\usepackage{mathabx,mathrsfs}
\usepackage{relsize}
\PassOptionsToPackage{hyphens}{url}\usepackage{hyperref}
\usepackage{xurl}
\usepackage[T1]{fontenc}

\usepackage[utf8]{inputenc}

\usepackage{microtype}

\usepackage{inconsolata}

\usepackage{natbib}

\usepackage{microtype}
\usepackage{pifont}
\usepackage{bm}
\usepackage{amsmath}
\usepackage{amssymb}
\usepackage{booktabs}
\usepackage{multirow}
\usepackage{float}
\usepackage{subcaption}
\usepackage{paralist}
\usepackage{tcolorbox}
\usepackage{scalerel}
\usepackage{tikz}
\usepackage{tikz-dependency}
\usepackage{tikz-qtree}
\usepackage{tikz-layers}
\usepackage{tikzsymbols}
\usepackage{pgfplots}
\usepackage{graphicx}
\usepackage{adjustbox}
\usepgflibrary{arrows.meta}
\usepgfplotslibrary{fillbetween, statistics, groupplots}
\usetikzlibrary{fit, calc, plotmarks, decorations.pathreplacing, decorations.markings, positioning, arrows.meta, shapes.geometric, backgrounds,graphs}
\pgfplotsset{compat=1.18}
\usepackage{xcolor,colortbl}

\def\pngemoji@insert#1{\scalerel*{\includegraphics{assets/#1.png}}{X}}
\newcommand\firstplace{\pngemoji@insert{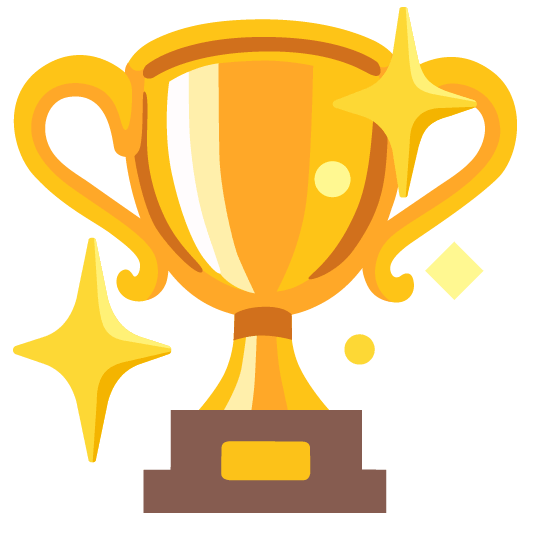}}
\newcommand\secondplace{\pngemoji@insert{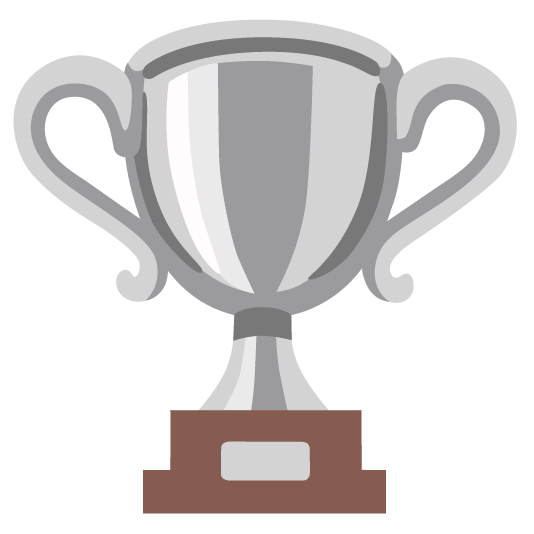}}
\newcommand\thirdplace{\pngemoji@insert{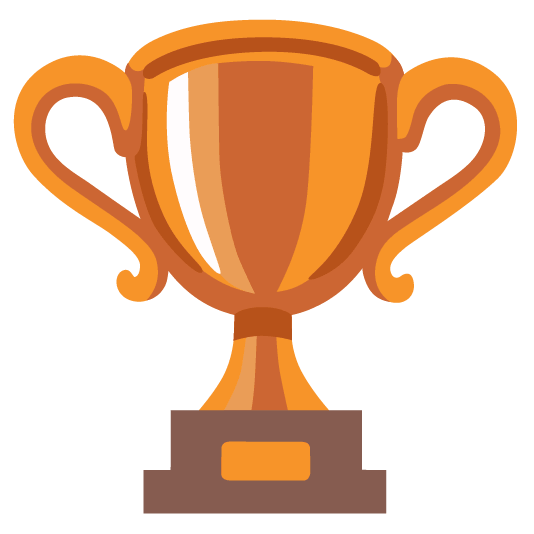}}

\definecolor{json_blue}{RGB}{15, 89, 164}
\definecolor{json_red}{RGB}{192, 44, 53}
\definecolor{figure_green}{RGB}{32, 137, 77}
\definecolor{figure_blue}{RGB}{23, 129, 181}
\definecolor{figure_red}{RGB}{192, 44, 53}
\definecolor{figure_orange}{RGB}{250, 126, 35}
\definecolor{table_blue}{RGB}{92, 179, 204}
\definecolor{table_red}{RGB}{238, 63, 77}
\definecolor{table_orange}{RGB}{250, 126, 35}
\newcommand\tab{\phantom{\qquad}}
\newcommand\jsonkey[1]{\texttt{\textbf{\textcolor{json_blue}{#1}}}}
\newcommand\mainstream[1]{\texttt{\textbf{\textcolor{json_blue}{#1}}}}
\newcommand\sideline[1]{\texttt{{\textcolor{gray}{#1}}}}
\newcommand\keyword[1]{\textbf{\textcolor{json_red}{#1}}}
\newcommand\depword[1]{\textcolor{figure_blue}{\textit{#1}}}
\newcommand\optionkey[1]{{\textbf{#1}}}

\newcommand\yes{\textcolor[RGB]{32, 127, 76}{\ding{51}}}%
\newcommand\no{\textcolor[RGB]{194, 31, 48}{\ding{55}}}%
\newcommand\wz{\phantom{0}}

\newcommand\matchany{\boldsymbol{\cdot}}
\newcommand\boldrecursivematch{\text{>$^{\kern-0.1em\text{*}}$}}
\newcommand\recursivematch{\boldrecursivematch}

\pgfdeclarelayer{background}  
\pgfdeclarelayer{above}       
\pgfsetlayers{background,main,above}  

\title{How Well Do Large Language Models Understand Syntax?\\An Evaluation by Asking Natural Language Questions}

\ifarxiv
    \author{
        Houquan Zhou$^1$, Yang Hou$^{1}$, Zhenghua Li$^1$, Xuebin Wang$^1$\\%
        {\bf Zhefeng Wang}$^2$, {\bf Xinyu Duan}$^2$, {\bf Min Zhang}$^1$,\\%
        $^1$ Institute of Artificial Intelligence, School of Computer Science and Technology,\\%
        Soochow University, China;\\%
        \texttt{\{hqzhou,yhou1,xbwang15\}@stu.suda.edu.cn, \{zhli13,minzhang\}@suda.edu.cn}\\%
        $^2$ Huawei Cloud, China\\%
        \texttt{\{wangzhefeng,duanxinyu\}@huawei.com}
    }
\else
    \author{Anonymous}
\fi

\begin{document}

\maketitle
\begin{abstract}
    While recent advancements in large language models (LLMs) bring us closer to achieving artificial general intelligence, the question persists: \textit{Do LLMs truly understand language, or do they merely mimic comprehension through pattern recognition?}
This study seeks to explore this question through the lens of syntax, a crucial component of sentence comprehension.
Adopting a natural language question-answering (Q\&A) scheme, we craft questions targeting nine syntactic knowledge points that are most closely related to sentence comprehension.
Experiments conducted on 24 LLMs suggest that most have a limited grasp of syntactic knowledge, exhibiting notable discrepancies across different syntactic knowledge points.
In particular, questions involving prepositional phrase attachment pose the greatest challenge, whereas those concerning adjectival modifier and indirect object are relatively easier for LLMs to handle.
Furthermore, a case study on the training dynamics of the LLMs reveals that the majority of syntactic knowledge is learned during the initial stages of training, hinting that simply increasing the number of training tokens may not be the `\textit{silver bullet}' for improving the comprehension ability of LLMs.

\end{abstract}

\section{Introduction}
The rapid advancement of large language models (LLMs) has showcased their impressive abilities.
Given a few exemplars or a set of instructions, LLMs can effectively handle a wide range of tasks, from traditional tasks like machine translation and summarization to more sophisticated, human-like activities such as solving mathematical problems, logical reasoning, and even planning.
Distinctly different from their predecessors, which often required fine-tuning for specific tasks, LLMs are viewed as a significant stride towards artificial general intelligence (AGI).

Yet, even as we are surprised by the prowess of LLMs, questions about their true understanding of language arise.
As black-boxs, do these models truly comprehend human language, or do they complete tasks by memorizing surface-level lexical patterns?
Do LLMs understand sentences based on syntactic rules, or do they treat language as merely \textit{a bag of words}?

Finding answers to these questions is of great importance to the LLM research community.
Consider human-centric evaluation benchmarks such as MMLU~\cite{hendrycks-etal-2021-measuring} and AGIEval~\cite{zhong-etal-2023-agieval}, which comprise questions intended for humans, presuming test-takers' competent language understanding, an assumption that may not hold true for LLMs.
Consequently, when an LLM errs in its response, discerning the root cause becomes convoluted.
The error could be a manifestation of the model's knowledge gaps, an inability to reason, or simply a failure to understand the question due to a lack of syntactic knowledge.
Measuring LLMs' syntactic knowledge is thus critical to understanding the true capabilities of LLMs.
\begin{figure}[tbp!]
    \captionsetup[subfigure]{skip=2pt}
    \centering
    \begin{minipage}[b]{0.95\linewidth}
        \small
        \begin{tcolorbox}[
                left=0pt,
                right=0pt,
                top=0pt,
                bottom=0pt,
                boxsep=6pt,
                colback=json_blue!3,
                colframe=black
            ]
            \jsonkey{Sentence}:\wz Pierre Vinken \textbf{\textit{will join}} the board as a nonexecutive director Nov. 29.\\[-3pt]
            \tikz \draw[densely dashed] (0, 0) -- (1.0\linewidth, 0); \\
            \jsonkey{Question}:\wz In the above sentence, the \textbf{\textit{grammatical subject}} of ``\textbf{will join}'' is \_\_\_\_\_\_\_\_\_\_\_\_. \\[1pt]
            \jsonkey{Options}: \\
            \tab \optionkey{A}. The board \\
            \tab \optionkey{B}. Pierre Vinken \\
            \tab \optionkey{C}. 61 years old \\
            \tab \optionkey{D}. A nonexecutive director \\[-3pt]
            \tikz \draw[densely dashed] (0, 0) -- (1.0\linewidth, 0); \\
            \jsonkey{Answer}: \textbf{B}
        \end{tcolorbox}
    \end{minipage}
    \caption{
        A example of syntactic knowledge question in the natural language question format.
    }
    \label{fig:question_example}
\end{figure}

To measure syntactic knowledge in LLMs, we must first determine \textbf{on which aspects of syntax we should focus}.
In contrast to prior work that focuses on aspects such as forming grammatically correct sentences, explaining specific syntactic phenomena \cite[\textit{inter alia}]{warstadt-etal-2019-neural,warstadt-etal-2020-blimp,gauthier-etal-2020-syntaxgym}, or depicting the hierarchical structure of sentences \cite[\textit{inter alia}]{hall-maudslay-etal-2020-tale,newman-etal-2021-refining, kim-etal-2023-reconstruction}, we concentrate on the comprehension aspect of syntax.
Therefore, our study emphasizes the syntactic knowledge of grammatical relations, which are more closely related to sentence understanding.
Evaluating the ability of a language model to identify subjects, objects, and other syntactic roles in a sentence.
Additionally, we also explore the ability of LLMs in resolving syntactic ambiguity.
In total, we select nine syntactic knowledge to evaluate, detailed in Table~\ref{tab:syntactic_knowledge}.

Then we turn to the methodology: \textbf{How should we evaluate syntactic knowledge in LLMs?}
Prior work has proposed two main approaches: probing and prompting.
However, both approaches have their limitations.
The probing approach requires access to hidden states, which are not available for API-only models like the GPT series, whereas the conventional prompting approach requires designing complex prompts and sophisticated decoding methods~\cite{roy-etal-2022-benchclamp}.
In response to these limitations, we utilizes a specific form of prompting, the natural language question-answering (Q\&A) paradigm.
This approach is an recently-mainstream and LLM-friendly evaluation method \cite{cobbe-etal-2021-training,hendrycks-etal-2021-measuring,zhong-etal-2023-agieval,huang-etal-2023-c-eval}.
For a thorough investigation, we design three question formats: True/False, Multiple Choice, and Fill in the Blank.
A example of the questions is depicted in Figure~\ref{fig:question_example}.

We conducted extensive experiments on 24 LLMs from 6 distinct families, including the state-of-the-art \texttt{GPT4}, the open-source \texttt{LLaMA\,1/2}, and other popular models, under both zero-shot and few-shot settings.
Our findings indicate that while most LLMs have a partial grip on syntactic knowledge, \texttt{GPT4} demonstrates exceptional superiority in all tested scenarios.
Closer examination showed that the prepositional phrase attachment (\texttt{PPA}) questions poses the greatest challenge, whereas adjectival modifier (\texttt{ADJ}) and indirect objects (\texttt{IO}) are comparatively simpler for LLMs to process.
Interestingly, we also observe that chat fine-tuning exhibits potential benefits for \texttt{PPA} difficulties.

Additionally, a case study on \texttt{Baichuan2} explores how syntactic knowledge evolves throughout training.
Our observations indicate that the majority of syntactic learning takes place in the early stages of training, suggesting that merely increasing the training tokens may not be the best way to improve syntactic knowledge.

In summary, our main contributions are:
\begin{asparaitem}[$\bullet$]
    \item We introduce a syntactic evaluation framework that evaluates LLMs' syntactic knowledge by asking LLMs natural language questions.
    \item Our comprehensive experiments across 24 LLMs reveals that most of LLMs are partially grasping syntactic knowledges.
    \item We also dip into the learning curve of syntactic knowledge and find that the majority of this knowledge is acquired during the initial stages.
\end{asparaitem}

We hope that our research is a step towards a more comprehensive understanding of LLMs' strengths and limitations.

\ifarxiv
    Our code is available at \url{https://github.com/Jacob-Zhou/SynEval}.
\else
    Our code and dataset will be publicly available at \url{https://github.com}.
\fi

\section{Design \& Construction of Evaluation Framework}
\label{sec:evaluation_framework}
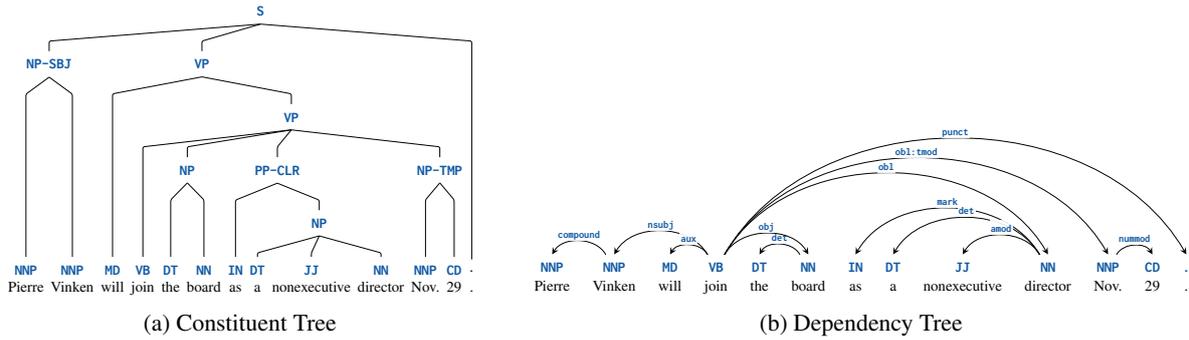
\begin{figure*}[tbp!]
    \captionsetup[subfigure]{skip=2pt}
    \centering
    \subfloat[Constituent Tree]{%
        \centering
        \label{fig:cfg_example}
        \scalebox{0.8}{%
            \begin{tikzpicture}[
                    level distance=25pt,
                    sibling distance=-4pt,
                    every tree node/.style={align=center,anchor=base,font=\scriptsize},
                    frontier/.style={distance from root=130pt},
                    edge from parent/.style={draw,edge from parent path={(\tikzparentnode.south) {[rounded corners=0.8pt]-- ($(\tikzchildnode |- \tikzparentnode.south) + (0, -8pt)$) -- (\tikzchildnode)}}},
                ]
                \Tree [.\jsonkey{S}
                    [.\jsonkey{NP-SBJ}
                        \jsonkey{NNP}\\Pierre \jsonkey{NNP}\\Vinken ]
                    [.\jsonkey{VP}
                        \jsonkey{MD}\\will
                        [.\jsonkey{VP}
                            \jsonkey{VB}\\join
                            [.\jsonkey{NP} \jsonkey{DT}\\the \jsonkey{NN}\\board ]
                            [.\jsonkey{PP-CLR}
                                \jsonkey{IN}\\as
                                [.\jsonkey{NP} \jsonkey{DT}\\a \jsonkey{JJ}\\nonexecutive \jsonkey{NN}\\director ] ]
                            [.\jsonkey{NP-TMP} \jsonkey{NNP}\\Nov. \jsonkey{CD}\\29 ] ] ] $\cdot$\\. ]
                \node[draw=none, inner sep=0pt] at (0, -4.8) {};
            \end{tikzpicture}
        }%
    }%
    \hfill
    \subfloat[Dependency Tree]{%
        \centering
        \label{fig:dg_example}
        \scalebox{0.8}{%
            \begin{dependency}[font=\scriptsize, edge vertical padding=-2pt, arc edge, arc angle=60, label style={above}, label style={fill=white,draw=none,inner sep=0.5pt,outer sep=0.5pt, rounded corners=0pt, fill opacity=0.9, text opacity=1}]
                \begin{deptext}[column sep=0.2cm]
                    \jsonkey{NNP} \& \jsonkey{NNP} \& \jsonkey{MD} \& \jsonkey{VB} \& \jsonkey{DT} \& \jsonkey{NN} \& \jsonkey{IN} \& \jsonkey{DT} \& \jsonkey{JJ} \& \jsonkey{NN} \& \jsonkey{NNP} \& \jsonkey{CD} \& \jsonkey{.} \\[-2pt]
                    Pierre \& Vinken \& will \& join \& the \& board \& as \& a \& nonexecutive \& director \& Nov. \& 29 \& . \\
                \end{deptext}
                \depedge{2}{1}{\jsonkey{compound}}
                \depedge{4}{2}{\jsonkey{nsubj}}
                \depedge{4}{3}{\jsonkey{aux}}
                \depedge{6}{5}{\jsonkey{det}}
                \depedge{4}{6}{\jsonkey{obj}}
                \depedge{4}{10}{\jsonkey{obl}}
                \depedge{10}{7}{\jsonkey{mark}}
                \depedge{10}{8}{\jsonkey{det}}
                \depedge{10}{9}{\jsonkey{amod}}
                \depedge{4}{11}{\jsonkey{obl:tmod}}
                \depedge{11}{12}{\jsonkey{nummod}}
                \depedge{4}{13}{\jsonkey{punct}}
            \end{dependency}
        }%
    }%
    \caption{
        Two types of syntactic trees for the same sentence.
    }
    \label{fig:syntax_examples}
\end{figure*}

In this study, we aim to investigate whether a LLM has essential syntax to understand a sentence.
To this end, we introduce a novel syntactic evaluation framework, in which we evaluate LLMs by asking them natural language questions tailored to uncover their syntactic knowledge of a sentence.

This section details the rationale behind our approach, outlines the core principles guiding our evaluation design, describes the process of crafting the syntactic questions, and discusses the methodology adopted in constructing the evaluation framework.

\subsection{Motivation}
\label{sec:design}
The primary objective of this evaluation is to find a way to investigate whether a language model has essential syntax to understand a sentence.

The syntax of a language is the consensus of how to arrange words to express specific meanings.
Only when words are arranged correctly can a sentence convey the writer's intended meaning.
Similarly, only when the reader understands the syntax can they fully grasp the sentence's meaning.
Therefore, the ability to understand a sentence is based on the syntactic knowledge of the reader.

\subsection{Design Principles}
\label{subsec:design_principles}
We follow two core principles in the design of our evaluation:
\paragraph{Relevance to understanding}
The first principle is that the syntactic knowledge we investigate in our evaluation should be directly related to the understanding of a sentence.
If a language model fails to identify this knowledge, it will probably fail to understand the sentence correctly.

\paragraph{Ease of Evaluation}
Our second principle is the simplicity of the evaluation process.
The notion for syntactic knowledges must be universal and easily comprehensible, thus precluding the necessity for specialized, academic, or domain-specific linguistic expertise.
Additionally, the evaluation methodology should avoid the need to access a model's hidden states, which is not available for API-only models such as the ChatGPT series.
Lastly, the evaluation should leverage the model's strength in generating natural language responses rather than demanding strict structural outputs, like bracketed parse trees or CoNLL-formatted structures.

\subsection{Selection of Syntactic Knowledge}

According to the Lexical-Functional Grammar theory, the syntactic structure of a sentence can be divided into two parts: constituent structure (c-structure) and functional structure (f-structure).
The c-structure provides a hierarchical framework, illustrating how individual components sequentially combine to form a complete sentence.
This can be analogized to a LEGO instruction manual for constructing a sentence.
For example, the noun phrase ``\textit{I}'' and the verb phrase ``\textit{am Batman}'' are combined to form the sentence ``\textit{I am Batman}''.
On the other hand, the f-structure is represented as a series of key-value pairs, detailing the functions of phrases and words, identifying, such as, which phrase serves as the subject and which as the object.
For example, in the sentence ``\textit{What I want is a car}'',
the f-structure is \texttt{Subject:} ``\textit{What I want}'' and \texttt{Object:} ``\textit{a car}''.

\begin{figure*}[tbp!]
    \captionsetup[subfigure]{skip=2pt}
    \centering
    {
        \hfill%
        \subfloat[]{%
            \centering
            \label{fig:gs_pattern}%
            \scalebox{0.8}{%
                \begin{tikzpicture}[
                        level distance=25pt,
                        every tree node/.style={align=center,anchor=base,font=\scriptsize, inner sep=1pt, outer sep=1pt},
                        frontier/.style={distance from root=45pt},
                        edge from parent/.style={draw,edge from parent path={(\tikzparentnode.south) {[rounded corners=0.8pt]-- ($(\tikzchildnode |- \tikzparentnode.south) + (0, -8pt)$) -- (\tikzchildnode)}}},
                    ]
                    \Tree [.\mainstream{S}
                        [.\sideline{$\matchany$*} ]
                        [.\mainstream{$\matchany$-SBJ} \edge[roof]; \texttt{\textbf{GS\strut}} ]
                        [.\sideline{$\matchany$*} ]
                        [.\mainstream{VP} ]
                        [.\sideline{$\matchany$*} ]
                    ]
                \end{tikzpicture}
            }%
        }%
        \hfill%
        \subfloat[]{%
            \centering
            \label{fig:sc_pattern}%
            \scalebox{0.8}{%
                \begin{tikzpicture}[
                        level distance=25pt,
                        every tree node/.style={align=center,anchor=base,font=\scriptsize, inner sep=1pt, outer sep=1pt},
                        frontier/.style={distance from root=85pt},
                        edge from parent/.style={draw,edge from parent path={(\tikzparentnode.south) {[rounded corners=0.8pt]-- ($(\tikzchildnode |- \tikzparentnode.south) + (0, -8pt)$) -- (\tikzchildnode)}}},
                    ]
                    \tikzset{level 1/.style={level distance=15pt}}
                    \Tree
                    [.\mainstream{\textasciitilde VP}
                    [.\node[fill=white](topv){\mainstream{VP$\boldrecursivematch$}};
                    [.\mainstream{(MD|TO|VB@)} \node (v0) {$v_0\mathstrut$}; ]
                    [.\sideline{(RB|ADVP)*} ]
                    \edge[densely dashed];
                    [.\mainstream{VP}
                    [.\mainstream{VB@}
                        \node (vn) {$v_n\mathstrut$};
                    ]
                    [.\sideline{(RB|ADVP)*} ]
                    [.\mainstream{$\matchany$-PRD} \edge[roof]; \texttt{\textbf{SC\strut}} ]
                    [.\sideline{$\matchany$*} ]
                    ]
                    [.\sideline{$\matchany$*} ]
                    ]
                    ]
                    \begin{scope}[on background layer]
                        \node (mvp) [draw, fill=gray!10, fit=(v0) (vn), inner sep=0pt, rounded corners=2pt, densely dashed] {};
                        \node [anchor=base, font=\scriptsize, inner sep=1pt, outer sep=1pt] at ($(v0.base)!0.5!(vn.base)$) {$\ldots\mathstrut$};
                        \node [anchor=base west, font=\scriptsize, inner sep=1pt, outer sep=1pt] at ($(vn.base east) + (0.2em, 0)$) {\texttt{\textbf{MVP}}};
                    \end{scope}
                \end{tikzpicture}
            }%
        }%
        \hfill%
        \subfloat[]{%
            \centering
            \label{fig:do_pattern}%
            \scalebox{0.8}{%
                \begin{tikzpicture}[
                        level distance=25pt,
                        every tree node/.style={align=center,anchor=base,font=\scriptsize, inner sep=1pt, outer sep=1pt},
                        frontier/.style={distance from root=85pt},
                        edge from parent/.style={draw,edge from parent path={(\tikzparentnode.south) {[rounded corners=0.8pt]-- ($(\tikzchildnode |- \tikzparentnode.south) + (0, -8pt)$) -- (\tikzchildnode)}}},
                    ]
                    \tikzset{level 1/.style={level distance=15pt}}
                    \Tree [.\mainstream{\textasciitilde VP}
                    [.\mainstream{VP$\boldrecursivematch$}
                    [.\mainstream{(MD|TO|VB@)} \node (v0) {$v_0\mathstrut$}; ]
                    [.\sideline{(RB|ADVP)*} ]
                    \edge[densely dashed];
                    [.\mainstream{VP}
                    [.\mainstream{VB@}
                        \node (vn) {$v_n\mathstrut$};
                    ]
                    [.\sideline{(RB|ADVP)*} ]
                    [.\mainstream{(NP|S|SBAR|SQ)} \edge[roof]; \texttt{\textbf{DO\strut}} ]
                    [.\sideline{(\textasciitilde (NP|S|SBAR|SQ))*} ]
                    ]
                    [.\sideline{$\matchany$*} ]
                    ]
                    ]
                    \begin{scope}[on background layer]
                        \node (mvp) [draw, fill=gray!10, fit=(v0) (vn), inner sep=0pt, rounded corners=2pt, densely dashed] {};
                        \node [anchor=base, font=\scriptsize, inner sep=1pt, outer sep=1pt] at ($(v0.base)!0.5!(vn.base)$) {$\ldots\mathstrut$};
                        \node [anchor=base west, font=\scriptsize, inner sep=1pt, outer sep=1pt] at ($(vn.base east) + (0.2em, 0)$) {\texttt{\textbf{MVP}}};
                    \end{scope}
                \end{tikzpicture}
            }%
        }%
    }
    \caption{
        Three examples of syntactic patterns.
        The pattern~\ref{fig:gs_pattern} is used to extract the grammatical subject (\texttt{GS}) of a sentence; the pattern~\ref{fig:sc_pattern} is used to extract the subject complement (\texttt{SC}) of a verb phrase (\texttt{MVP}); the pattern~\ref{fig:do_pattern} is used to extract the direct object (\texttt{DO}) of a verb phrase.
        ``\texttt{$\matchany$}'' matches any pharse or word; ``\texttt{*}'' matches zero or more times horizontally;
        ``\texttt{$\recursivematch$}'' matches zero or more times recursively; ``\texttt{|}'' matches either the left or the right pattern; ``\texttt{\textasciitilde}'' is the negation of the pattern;
        ``\texttt{VB@}'' matches verb related part-of-speech tags, such as ``\texttt{VB}'', ``\texttt{VBZ}''.
    }
    \label{fig:patterns}
\end{figure*}
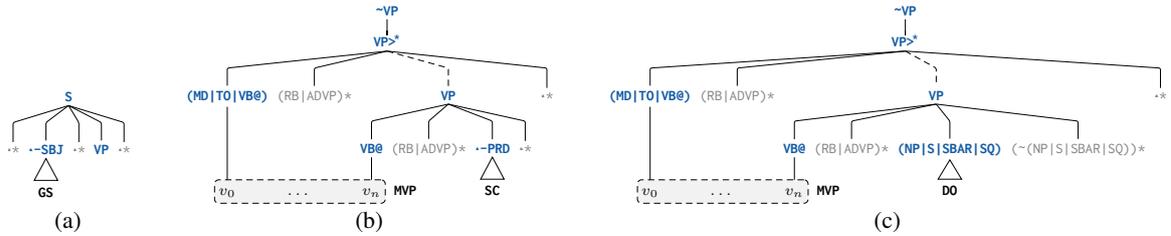

Recall that our objective is to investigate whether a language model can use syntactic knowledge to identify the elements of a sentence in order to understand it, rather than to generate a syntactically correct sentence, which has been extensively studied in previous work \cite{warstadt-etal-2019-neural,warstadt-etal-2020-blimp,gauthier-etal-2020-syntaxgym}.
Therefore, we mainly focus on the f-structure.
That is, we want to know whether a language model can identify the subject, object, and other syntactic elements of a sentence.
Except for the f-structure, we also interested in two important syntactic knowledge of s-structure: the coordination structure and the prepositional phrase attachment, which have a close relationship with the syntactic ambiguity problem.

The full list of syntactic knowledge we investigate is shown in Table~\ref{tab:syntactic_knowledge}.

\subsection{Selection of Paradigm}
In line with the second design principle, we follow the recent mainstream approach of LLMs evaluating work, such as GSM8k \cite{cobbe-etal-2021-training}, MMLU \cite{hendrycks-etal-2021-measuring}, and AGIEval \cite{zhong-etal-2023-agieval}, using a question-answering (Q\&A) paradigm.
That is, we pose a natural language question to the model as a prompt, and the model is expected to answer the question in natural language as well.
We include three question types: True / False, Multiple Choice, and Fill in the Blank, for a holistic evaluation.

\subsection{Design of Questions}
\label{sec:design_questions}
In the design of our questions, we adopted traditional syntactic concepts to guide our investigation into syntactic knowledge.
The questions are structured such that the answers are phrases or full words from the sentence, mirroring the more natural human approach to responding to questions, rather than just the head word of the phrase.
For example, as illustrated in Figure~\ref{fig:dg_example}, when asked, ``\textit{What is the prepositional object of `as'?}'', most individuals are tended to answer with the complete phrase ``\textit{a nonexecutive director},'' as opposed to the singular head word ``\textit{director}.''\\
An instance of the questions is shown in Figure~\ref{fig:question_example}.
\begin{table*}[t!]
    \setlength{\tabcolsep}{7.5pt}
    \centering
    \scalebox{0.9}{%
        \begin{tabular}{lc|ccc|l}
            \toprule
            \textbf{Syntactic Knowledge Points}                                         & \textbf{Abbr.}                         & \textbf{\#TF}              & \textbf{\#MC}        & \textbf{\#FITB}      & \textbf{Example}                                                    \\
            \midrule
            \textbf{G}rammatical \textbf{S}ubject                                       & \textbf{\texttt{GS}}                   & 130                        & 105                  & 105                  & \keyword{Desks} \depword{are cleared} by John.                      \\
            \textbf{S}ubject \textbf{C}omplement                                        & \textbf{\texttt{SC}}                   & 130                        & \wz85                & \wz85                & John \depword{is} \keyword{a teacher}.                              \\
            \textbf{D}irect \textbf{O}bject                                             & \textbf{\texttt{DO}}                   & 150                        & 145                  & 145                  & John \depword{gave} me \keyword{a book}.                            \\
            \textbf{I}ndirect \textbf{O}bject                                           & \textbf{\texttt{IO}}                   & \wz30                      & \wz20                & \wz20                & John \depword{gave} \keyword{me} a book.                            \\
            \textbf{M}ain \textbf{V}erb \textbf{P}hrase                                 & \textbf{\texttt{MVP}}                  & 440$^\ddagger$\kern-0.45em & 170                  & 170                  & John \keyword{gave} me a book.                                      \\
            \textbf{ADJ}ectival modifier$^\dagger$                                      & \textbf{\texttt{ADJ}}                  & 185                        & 165                  & 135                  & I enjoy \depword{the book} \keyword{John gave me}.                  \\
            \textbf{ADV}erbial modifier (Adjunct)                                       & \textbf{\texttt{ADV}}                  & 165                        & 125                  & 115                  & I \depword{read} the book \keyword{quickly}.                        \\
            \midrule
            \textbf{CO}ordination                                                       & \textbf{\texttt{CO}}                   & 165                        & 160                  & 155                  & We \keyword{will play} football \underline{and} \keyword{watch} TV. \\
            \multirow{2}{*}{\textbf{P}repositional \textbf{P}hrase \textbf{A}ttachment} & \multirow{2}{*}{\textbf{\texttt{PPA}}} & \multirow{2}{*}{110}       & \multirow{2}{*}{100} & \multirow{2}{*}{100} & I like \keyword{the book} \depword{on my shelf}.                    \\
                                                                                        &                                        &                            &                      &                      & I \keyword{hide} the book \depword{on my shelf}.                    \\
            \bottomrule
        \end{tabular}
    }
    \caption{Syntactic knowledge points in our evaluation. $^\dagger$: We only consider post-modifier, such as relative clause and reduced relative clause in this work. $^\ddagger$: The questions of main verb phrase in True/False are the same as those in surface subject, subject complement, direct object, and indirect object, so we directly reuse the questions of these four syntactic knowledge points and do not count them in the total number of questions.}
    \label{tab:syntactic_knowledge}
\end{table*}

\subsection{Construction}
\label{sec:construction}
Instead of manually creating questions and answers, we propose to take advantage of existing syntactic annotations to automatically generate questions and answers.
In this subsection, we briefly introduce the automatic syntactic information extraction and question generation process.

\paragraph{Extracting Syntactic Information}
In this work, we extract syntactic information from the Penn Treebank (PTB), which is a widely used constituency treebank.
A example of the constituency tree is shown in Figure~\ref{fig:cfg_example}.

Why do we use constituency trees instead of dependency trees?
Extracting syntactic information from a sentence based on its dependency tree, as shown in Figure~\ref{fig:dg_example}, where the relationship between words is explicitly annotated, might seem more straightforward.
However, two main reasons prevent us from directly utilizing the dependency tree.
Firstly, most existing dependency treebanks have been automatically converted from constituency treebanks.
This conversion might introduce errors that we are unaware of.
Secondly, the dependency tree models the relationships between word pairs, making the extraction of answer phrases or full words difficult.

To extract syntactic information, we first learn the PTB guidelines carefully, figure out how the syntactic information is annotated, and design patterns for each type of syntactic knowledge.
Some of the patterns we design are shown in Figure~\ref{fig:patterns}.

Then, by searching for the patterns in the constituency tree, we extract the syntactic information of a sentence.
For example, the pattern shown in Figure~\ref{fig:gs_pattern} matches the ``\texttt{S}'' node that has both a immediate child labeled with ``\texttt{-SBJ}'' function tag and a immediate ``\texttt{VP}'' child.
We can then extract the immediate child with the ``\texttt{-SBJ}'' function tag as the subject of the sentence ``\texttt{S}''.

To ensure the quality of the extraction, the outliers that do not match any of the patterns are ignored.

\paragraph{Question Generation}
We manually design templates for each type of questions and each syntactic knowledge point.
Then, we use the extracted syntactic information to fill in the templates to generate questions.
Along with the question, we also generate the meta information, such as the syntactic category (e.g., noun phrase, \textit{that}-clause, etc.) of the answer or the words that fill in the placeholder, for the convenience of the future analysis.

\section{Experiments}
\label{sec:experiments}
\subsection{Experimental Setup}
Our experiments are conducted under two distinct settings: \textbf{Zero-shot} and \textbf{Few-shot}.
In both settings, the models are prompted to provide direct answers (referred to as the answer-only approach), \textbf{without} leveraging the {Chain of Thought} (CoT) technique\footnote{Due to the space limitations, we provide a detailed discussion of the CoT technique in Appendix~\ref{sec:appendix:cot}.}.

\subsection{Question Sampling}
After question creation, we collected 3,538,818 questions.
The detailed statistics of the questions are shown in Appendix~\ref{sec:appendix:original_question_dist}.
From the statistics, we observe that the number of questions for each syntactic knowledge point is extremely unbalanced.
For example, the number of questions for the knowledge point of main verb phrase (\texttt{MVP}) is 248 times that of the knowledge point of indirect object (\texttt{IO}).
Therefore, we conduct a balanced down-sampling to ensure that each syntactic knowledge point has a similar number of questions.

Specifically, we first combine the question type, the syntactic knowledge point, and the syntactic category of the answer into a tuple.
For each tuple, we randomly sample $k=5$ questions from those associated with it to form the evaluation set.
At the conclusion of this process, our test set comprises 3,170 questions, with detailed statistics presented in Table~\ref{tab:syntactic_knowledge}.
Employing a similar approach but with a reduced sample size, we derived an exemplar set containing 1,300 questions.

\begin{table*}[tbp!]
    \centering
    \scalebox{0.85}{%
        \begin{tabular}{lcccccccccccc}
            \toprule
                                                                       & \multicolumn{5}{c}{\textbf{Zero-shot}}      &                                             & \multicolumn{5}{c}{\textbf{Few-shot}}                                                                                                                                                                                                                                                                                                                                                                                                                  \\
            \cmidrule(l{0pt}r{0pt}){2-6}\cmidrule(l{0pt}r{0pt}){8-12}  & \textbf{TF}                                 & \textbf{MC}                                 & \multicolumn{2}{c}{\textbf{FITB}}           & \multirow{2}{*}{\textbf{OA}}                &                                             & \textbf{TF}                        & \textbf{MC}                                 & \multicolumn{2}{c}{\textbf{FITB}}           & \multirow{2}{*}{\textbf{OA}}                                                                                                                                                \\
            \cmidrule(l{0pt}r{0pt}){4-5}\cmidrule(l{0pt}r{0pt}){10-11} & Acc.                                        & Acc.                                        & Acc.                                        & $F_1$                                       &                                             &                                    & Acc.                                        & Acc.                                        & Acc.                                        & $F_1$                                       &                                             &                                   \\
            \midrule
            \texttt{Random}                                            & 50.00                                       & 25.00                                       & 0.68                                        & 23.21                                       & 28.66                                       &                                    & 50.00                                       & 25.00                                       & 0.68                                        & 23.21                                       & 28.66                                       &                                   \\
            \midrule
            \texttt{Mistral 7B}                                        & \cellcolor{table_blue!2.430}51.08           & \cellcolor{table_blue!35.197}50.42          & \cellcolor{table_blue!47.846}40.19          & \cellcolor{table_blue!45.546}57.01          & \cellcolor{table_blue!31.052}50.03          &                                    & \cellcolor{table_blue!14.618}56.50          & \cellcolor{table_blue!43.740}56.59          & \cellcolor{table_blue!66.503}55.60          & \cellcolor{table_blue!62.484}69.58          & \cellcolor{table_blue!43.626}58.56          &                                   \\
            \texttt{Mistral 7B\,(Instruct)}                            & \cellcolor{table_blue!17.212}57.65          & \cellcolor{table_blue!38.672}52.93          & \cellcolor{table_blue!42.917}36.12          & \cellcolor{table_blue!40.371}53.17          & \cellcolor{table_blue!33.572}51.74          &                                    & \cellcolor{table_blue!13.628}56.06          & \cellcolor{table_blue!40.989}54.60          & \cellcolor{table_blue!54.946}46.05          & \cellcolor{table_blue!53.191}62.68          & \cellcolor{table_blue!38.392}55.01          &                                   \\
            \midrule
            \texttt{Baichuan2 13B}                                     & \cellcolor{table_blue!4.747}52.11           & \cellcolor{table_blue!41.511}54.98          & \cellcolor{table_blue!43.026}36.21          & \cellcolor{table_blue!41.274}53.84          & \cellcolor{table_blue!32.043}50.71          &                                    & \cellcolor{table_blue!4.613}52.05           & \cellcolor{table_blue!45.240}57.67          & \cellcolor{table_blue!62.858}52.59          & \cellcolor{table_blue!58.185}66.39          & \cellcolor{table_blue!40.448}56.40          &                                   \\
            \texttt{Baichuan2 13B\,(Chat)}                             & \cellcolor{table_blue!21.443}59.53          & \cellcolor{table_blue!42.798}55.91          & \cellcolor{table_blue!31.389}26.60          & \cellcolor{table_blue!30.777}46.05          & \cellcolor{table_blue!31.871}50.59          &                                    & \cellcolor{table_blue!16.028}57.12          & \cellcolor{table_blue!44.940}57.46          & \cellcolor{table_blue!53.295}44.69          & \cellcolor{table_blue!50.693}60.83          & \cellcolor{table_blue!39.528}55.78          &                                   \\
            \midrule
            \texttt{Falcon 40B}                                        & \cellcolor{table_blue!6.030}52.68           & \cellcolor{table_blue!32.622}48.56          & \cellcolor{table_blue!32.563}27.57          & \cellcolor{table_blue!29.510}45.11          & \cellcolor{table_blue!24.897}45.86          &                                    & \cellcolor{table_blue!17.212}57.65          & \cellcolor{table_blue!40.472}54.23          & \cellcolor{table_blue!55.297}46.34          & \cellcolor{table_blue!52.364}62.07          & \cellcolor{table_blue!38.912}55.36          &                                   \\
            \texttt{Falcon 40B\,(Instruct)}                            & \cellcolor{table_blue!18.068}58.03          & \cellcolor{table_blue!32.358}48.37          & \cellcolor{table_blue!34.561}29.22          & \cellcolor{table_blue!30.238}45.65          & \cellcolor{table_blue!27.972}47.95          &                                    & \cellcolor{table_blue!12.990}55.77          & \cellcolor{table_blue!39.748}53.71          & \cellcolor{table_blue!55.144}46.22          & \cellcolor{table_blue!52.795}62.39          & \cellcolor{table_blue!37.779}54.59          &                                   \\
            \midrule
            \texttt{Llama 65B}                                         & \cellcolor{table_blue!19.328}58.59          & \cellcolor{table_blue!42.923}56.00          & \cellcolor{table_blue!54.434}45.63          & \cellcolor{table_blue!53.105}62.62          & \cellcolor{table_blue!40.204}56.24          &                                    & \cellcolor{table_blue!5.033}52.24           & \cellcolor{table_blue!41.852}55.23          & \cellcolor{table_blue!73.171}61.10          & \cellcolor{table_blue!68.593}74.11          & \cellcolor{table_blue!43.329}58.36          &                                   \\
            \midrule
            \texttt{Llama2 70B}                                        & \cellcolor{table_blue!15.953}57.09          & \cellcolor{table_blue!56.963}66.14          & \cellcolor{table_blue!55.136}46.21          & \cellcolor{table_blue!54.385}63.57          & \cellcolor{table_blue!44.828}59.37          &                                    & \cellcolor{table_blue!16.515}57.34          & \cellcolor{table_blue!58.080}66.95          & \cellcolor{table_blue!73.757}61.59          & \cellcolor{table_blue!69.936}75.11          & \cellcolor{table_blue!51.964}64.21          &                                   \\
            \texttt{Llama2 70B\,(Chat)}                                & \cellcolor{table_blue!15.750}57.00          & \cellcolor{table_blue!50.649}61.58          & \cellcolor{table_blue!50.437}42.33          & \cellcolor{table_blue!49.979}60.30          & \cellcolor{table_blue!40.784}56.63          &                                    & \cellcolor{table_blue!22.710}60.09          & \cellcolor{table_blue!60.438}68.65          & \cellcolor{table_blue!66.822}55.86          & \cellcolor{table_blue!63.899}70.63          & \cellcolor{table_blue!51.646}64.00          &                                   \\
            \midrule
            \texttt{GPT3.5}                                            & \cellcolor{table_blue!21.443}59.53          & \cellcolor{table_blue!46.662}58.70          & \cellcolor{table_blue!66.192}55.34          & \cellcolor{table_blue!64.990}71.44          & \cellcolor{table_blue!46.549}60.54          &                                    & \cellcolor{table_blue!30.105}63.38          & \cellcolor{table_blue!67.265}73.58          & \cellcolor{table_blue!68.541}57.28          & \cellcolor{table_blue!66.230}72.36          & \cellcolor{table_blue!56.460}67.26          & \kern-0.5em\thirdplace\kern-0.5em \\
            \texttt{GPT4}                                              & \cellcolor{table_blue!71.730}\textbf{81.88} & \cellcolor{table_blue!87.494}\textbf{88.19} & \cellcolor{table_blue!76.655}\textbf{63.98} & \cellcolor{table_blue!73.533}\textbf{77.78} & \cellcolor{table_blue!75.718}\textbf{80.32} & \kern-0.5em\secondplace\kern-0.5em & \cellcolor{table_blue!87.367}\textbf{88.83} & \cellcolor{table_blue!93.157}\textbf{92.28} & \cellcolor{table_blue!83.122}\textbf{69.32} & \cellcolor{table_blue!80.702}\textbf{83.10} & \cellcolor{table_blue!83.766}\textbf{85.77} & \kern-0.5em\firstplace\kern-0.5em \\
            \bottomrule
        \end{tabular}
    }
    \caption{Main results of our evaluation.}
    \label{tab:main_results}
\end{table*}

\subsection{Evaluation Metrics}
For True/False and Multiple Choice questions, we employ standard accuracy, adhering to conventions set by previous work.
For Fill in the Blank questions, we utilize Accuracy (Acc.) and $F_1$ score (Question-wise averaging) as evaluation metrics.
Notably, compared to prior studies, we adopt a stricter $F_1$ score, in which we require that words in the predicted answer align in the same order as those in the ground truth answer.
To mitigate any potential issues arising from tokenization and punctuation discrepancies, we employ \texttt{NLTK} to re-tokenize the sentences and discard all punctuation before computing the Acc. and $F_1$ score.
Finally, we report the overall performance (OA) as the average of the three question types:

\vspace{-1em}
{\footnotesize
    \begin{equation}
        \textbf{OA} = \frac{1}{3}\!\left(\textbf{TF}_{Acc.}\!+\!\textbf{MC}_{Acc.}\!+\!\frac{1}{2}\!\left(\textbf{FITB}_{Acc.}\!+\!\textbf{FITB}_{F_1}\right)\right)
    \end{equation}
}

\subsection{Selection of Large Language Models}
We conduct comprehensive experiments on 24 large language models from 6 different families.
As shown in Table~\ref{tab:models}, the six families are as follows:
\begin{inparaenum}[1)]
    \item \textbf{Mistral},
    \item \textbf{Baichuan2},
    \item \textbf{Falcon},
    \item \textbf{LLaMA},
    \item \textbf{LLaMA2}, and
    \item \textbf{ChatGPT} series.
\end{inparaenum}
The detailed information about the models can be found in Appendix~\ref{sec:appendix:model_details}.

\section{Results and Findings}
In this section, we present the experimental results for LLMs and provide a series of findings based on the results.
We then conduct a case study on \texttt{Baichuan2}, of which the intermediate checkpoint is available to the public, to further investigate the relationship between the number of training tokens and the model's performance.
Due to space limitations, the detailed results of all models are presented in Appendix~\ref{sec:appendix:detailed_results}.

\subsection{Main Results}
\label{sec:main_findings}
The main results are shown in Table~\ref{tab:main_results}; the overall accuracy (OA) across different knowledge points is presented in Figure~\ref{tab:acc_vs_kp}; and the relationship between the number of parameters and the model's performance is shown in Figure~\ref{fig:acc_vs_params}.
From the results, we can observe several interesting findings:
\begin{asparaenum}[\bfseries I)]
    \item \textbf{LLMs is partially grasping syntax: }
    As shown in Table~\ref{tab:main_results} and \ref{tab:main_results_all}, the overall accuracy (OA) of all models larger than 1B is significantly higher than the random baseline, which indicates that LLMs do have the basic ability to understand syntax.
    However, only two models, \texttt{GPT4} and \texttt{GPT3.5}, have an OA greater than 60 in both zero-shot and few-shot settings, and only two other models, \texttt{Llama2\,70B} and \texttt{Llama2\,70B\,(Chat)}, whose OA is higher than 60 on few-shot setting.
    This indicates that most LLMs can not answer the syntactic knowledge questions very well, and there is still a long way to go.
    \item \textbf{Few-shot outperforms Zero-shot in most cases: }
    The zero-shot setting requires the model to understand the meaning of syntactic terms, such as ``\textit{subject}'' and ``\textit{object}'', and to identify the corresponding syntactic elements in the sentence.
    It is more difficult than the few-shot setting.
    As expected, compared to the few-shot setting, the zero-shot setting has a lower OA (from -2.88 to -11.42) on all models.
    However, there is one exception where some Chat/Instruct models have a higher True/False accuracy on zero-shot setting than few-shot setting.
    \item \textbf{GPT4 shows superior performance: }
    All results consistently show that \texttt{GPT4} outperforms other models by a large margin with an OA difference of 20.06 on zero-shot setting and 18.65 on few-shot setting.
    Even its results on the zero-shot setting are better than those of all other models in the few-shot setting.
    When we look at the results of different knowledge points, we can find that \texttt{GPT4} exceeds 85 OA on 7 out of 9 knowledge points on few-shot setting, among which the OA of indirect object (\texttt{IO}) are even higher than 95.
    Despite the superiority of \texttt{GPT4}, there are still some other models that outperform \texttt{GPT4} on some knowledge points.
    For example, when answering fill in the blank questions, \texttt{Llama2\,70B} outperforms \texttt{GPT4} on the knowledge point of adverbial modifier (\texttt{ADV}) on both zero-shot and few-shot settings and coordination (\texttt{CO}) on zero-shot setting.
    \begin{table*}[tb]
    \setlength{\tabcolsep}{6pt}
    \centering
    \scalebox{0.85}{%
        \begin{tabular}{lccccccccc}
            \toprule
            \textbf{Models}                 & \textbf{GS}                                 & \textbf{SC}                                 & \textbf{DO}                                 & \textbf{IO}                                 & \textbf{MVP}                                & \textbf{ADJ}                                & \textbf{ADV}                                & \textbf{PPA}                                & \textbf{CO}                                 \\
            \midrule
            \texttt{Mistral 7B}             & \cellcolor{table_blue!33.048}62.81          & \cellcolor{table_blue!27.828}57.68          & \cellcolor{table_blue!32.422}63.22          & \cellcolor{table_blue!38.099}68.76          & \cellcolor{table_blue!30.500}59.66          & \cellcolor{table_blue!34.070}64.06          & \cellcolor{table_blue!25.116}55.13          & \cellcolor{table_blue!9.480}38.26           & \cellcolor{table_blue!31.108}60.74          \\
            \texttt{Mistral 7B\,(Instruct)} & \cellcolor{table_blue!32.171}61.89          & \cellcolor{table_blue!23.120}52.80          & \cellcolor{table_blue!29.981}60.76          & \cellcolor{table_blue!24.968}55.42          & \cellcolor{table_blue!24.629}53.42          & \cellcolor{table_blue!28.726}58.51          & \cellcolor{table_blue!28.085}58.19          & \cellcolor{table_blue!4.593}33.16           & \cellcolor{table_blue!26.786}56.21          \\
            \midrule
            \texttt{Baichuan2 13B}          & \cellcolor{table_blue!29.989}59.61          & \cellcolor{table_blue!28.780}58.66          & \cellcolor{table_blue!30.622}61.41          & \cellcolor{table_blue!37.258}67.90          & \cellcolor{table_blue!31.461}60.68          & \cellcolor{table_blue!32.234}62.15          & \cellcolor{table_blue!24.393}54.39          & \cellcolor{table_blue!1.994}30.45           & \cellcolor{table_blue!26.548}55.96          \\
            \texttt{Baichuan2 13B\,(Chat)}  & \cellcolor{table_blue!34.005}63.81          & \cellcolor{table_blue!28.644}58.52          & \cellcolor{table_blue!29.198}59.97          & \cellcolor{table_blue!34.439}65.04          & \cellcolor{table_blue!28.294}57.31          & \cellcolor{table_blue!31.879}61.78          & \cellcolor{table_blue!20.897}50.79          & \cellcolor{table_blue!4.561}33.13           & \cellcolor{table_blue!26.630}56.05          \\
            \midrule
            \texttt{Falcon 40B}             & \cellcolor{table_blue!31.686}61.38          & \cellcolor{table_blue!25.676}55.45          & \cellcolor{table_blue!26.451}57.21          & \cellcolor{table_blue!34.305}64.90          & \cellcolor{table_blue!31.160}60.36          & \cellcolor{table_blue!30.265}60.11          & \cellcolor{table_blue!20.472}50.36          & \cellcolor{table_blue!7.360}36.05           & \cellcolor{table_blue!26.303}55.71          \\
            \texttt{Falcon 40B\,(Instruct)} & \cellcolor{table_blue!27.973}57.50          & \cellcolor{table_blue!26.371}56.17          & \cellcolor{table_blue!26.646}57.40          & \cellcolor{table_blue!27.768}58.26          & \cellcolor{table_blue!31.556}60.78          & \cellcolor{table_blue!32.619}62.55          & \cellcolor{table_blue!19.679}49.54          & \cellcolor{table_blue!7.840}36.55           & \cellcolor{table_blue!22.446}51.67          \\
            \midrule
            \texttt{Llama 65B}              & \cellcolor{table_blue!33.636}63.42          & \cellcolor{table_blue!30.634}60.58          & \cellcolor{table_blue!31.871}62.67          & \cellcolor{table_blue!40.652}71.35          & \cellcolor{table_blue!30.199}59.34          & \cellcolor{table_blue!35.339}65.38          & \cellcolor{table_blue!26.108}56.15          & \cellcolor{table_blue!11.005}39.86          & \cellcolor{table_blue!25.890}55.27          \\
            \midrule
            \texttt{Llama2 70B}             & \cellcolor{table_blue!40.717}70.83          & \cellcolor{table_blue!35.549}65.67          & \cellcolor{table_blue!32.557}63.36          & \cellcolor{table_blue!51.331}82.20          & \cellcolor{table_blue!36.081}65.59          & \cellcolor{table_blue!44.194}74.58          & \cellcolor{table_blue!31.616}61.82          & \cellcolor{table_blue!15.490}44.54          & \cellcolor{table_blue!31.048}60.68          \\
            \texttt{Llama2 70B\,(Chat)}     & \cellcolor{table_blue!38.832}68.86          & \cellcolor{table_blue!26.423}56.22          & \cellcolor{table_blue!36.315}67.14          & \cellcolor{table_blue!38.099}68.76          & \cellcolor{table_blue!40.569}70.36          & \cellcolor{table_blue!44.635}75.04          & \cellcolor{table_blue!29.852}60.00          & \cellcolor{table_blue!20.457}49.72          & \cellcolor{table_blue!26.899}56.33          \\
            \midrule
            \texttt{GPT3.5}                 & \cellcolor{table_blue!45.614}75.95          & \cellcolor{table_blue!39.656}69.93          & \cellcolor{table_blue!39.701}70.55          & \cellcolor{table_blue!49.580}80.42          & \cellcolor{table_blue!40.175}69.94          & \cellcolor{table_blue!40.341}70.57          & \cellcolor{table_blue!32.750}62.98          & \cellcolor{table_blue!29.292}58.94          & \cellcolor{table_blue!29.167}58.71          \\
            \texttt{GPT4}                   & \cellcolor{table_blue!58.798}\textbf{89.74} & \cellcolor{table_blue!55.850}\textbf{86.70} & \cellcolor{table_blue!56.022}\textbf{86.99} & \cellcolor{table_blue!65.579}\textbf{96.67} & \cellcolor{table_blue!54.624}\textbf{85.29} & \cellcolor{table_blue!61.381}\textbf{92.44} & \cellcolor{table_blue!43.024}\textbf{73.55} & \cellcolor{table_blue!50.901}\textbf{81.50} & \cellcolor{table_blue!56.779}\textbf{87.63} \\
            \midrule
            Avg.                            & \cellcolor{table_orange!26.010}55.44        & \cellcolor{table_orange!23.873}53.58        & \cellcolor{table_orange!24.488}55.23        & \cellcolor{table_orange!27.850}58.35        & \cellcolor{table_orange!25.176}54.00        & \cellcolor{table_orange!28.752}58.53        & \cellcolor{table_orange!19.779}49.65        & \cellcolor{table_orange!7.724}36.43         & \cellcolor{table_orange!22.403}51.62        \\
            \bottomrule
        \end{tabular}
    }
    \caption{Overall performance of each model under Few-shot setting at the knowledge point level.}
    \label{tab:acc_vs_kp}
\end{table*}
    \item \textbf{Parameter size impacts performance differently: }
    The relationship between parameter size and model performance is depicted in Figure~\ref{fig:acc_vs_params}.
    Within individual families, there's a general trend that aligns performance with parameter size: larger models tend to achieve better results.
    However, when comparing across different families, this correlation is not always consistent.
    For instance, the ``\texttt{Baichuan2\,7B}'' model outperforms all 13B models in Multiple Choice questions.
    \item \textbf{\texttt{PPA} tops difficulty, \texttt{ADJ} and \texttt{IO} rank as easiest: }
    Table~\ref{tab:acc_vs_kp} offers a granular analysis of results across different syntactic knowledge points.
    From the average results across all models, we can observe that the knowledge point of prepositional phrase attachment (\texttt{PPA}) is the most difficult one, with an OA of 36.43, while that of adjectival modifier (\texttt{ADJ}) and indirect object (\texttt{IO}) are the easiest ones, with an OA of 58.53 and 58.35, respectively.
    \item \textbf{Chat fine-tuning benefits \texttt{PPA} performance: }
    \label{finding:chat_ppa}
    From Table~\ref{tab:acc_vs_kp}, we can observe that the most of Chat/Instruct models have a higher OA on \texttt{PPA} than their foundation models.
    For example, the OA of \texttt{Llama2\,70B\,(Chat)} on \texttt{PPA} is 5.18 higher than that of \texttt{Llama2\,70B}, while inferior on almost all other knowledge points.
    The same phenomenon also appears on \texttt{Baichuan2\,13B} and \texttt{Baichuan2\,13B\,(Chat)}.
    We suggest that this is because that the correct understanding of \texttt{PPA} is crucial for the chat task.
    \item \textbf{Inconsistent knowledge generalize across question types: }
    When we compare the metrics of different question types, we can find that the knowledge does not generalize well across different question types.
    First, we observe that when the model has a high performance on one question type, it does not mean that it will also have a high performance on other question types.
    For example, as shown in Table~\ref{tab:main_results}, even when \texttt{Baichuan2\,13B} has outperformed random baseline by a large margin on Fill in the Blank questions, in which the model is required to generate the text of the answer, its OA on True/False questions is merely 2.05 higher than the random baseline.
    Second, we observe that the correlation between the performance on different question types is not consistent.
    The Kendall's $\tau$ and Pearson's $r$ correlation coefficients are shown in Table~\ref{tab:correlation} in Appendix~\ref{sec:appendix:detailed_results}.
    The results indicate that correlations between the performance on True/False and other question types are all lower than 0.8, meaning that there is no strong correlation between the performance on True/False and other question types.
    A typical example is that the Chat/Instruct versions have a higher accuracy on True/False and multiple choice questions than its foundation versions, but a lower accuracy on Fill in the Blank.
\end{asparaenum}%
\input{figures/acc_vs_params}
\input{figures/acc_vs_steps}
\subsection{Training Dynamics for Knowledge Points: A Case Study on \texttt{Baichuan2}}
We conduct a case study on \texttt{Baichuan2\,7B}\footnote{The \texttt{Baichuan2\,7B} model is one of the few models that publicly release intermediate checkpoints, which facilitates our case study.} to further explore the relationship between the pre-training process and the model's performance.
The results are presented in Figure~\ref{fig:acc_vs_tokens}.

The \texttt{Baichuan2} series has been trained with a total of 2.64T tokens.
Intermediate checkpoints were made available after every 220B tokens trained.

The results reveal several trends common to most knowledge points:
\begin{inparaenum}[1)]
    \item There is a positive correlation between the number of training tokens and performance across most knowledge points: the greater the number of tokens, the better the performance.
    \item After the initial training with 220B tokens, the model significantly exceeds the random baseline across most knowledge points, with improvements ranging from +7.18 to +18.26, except for \texttt{PPA}.
    \item The most substantial performance gains occur during the first 1.32T tokens; beyond this point, the improvements are considerably smaller across most knowledge points (average improvement of 2.88 vs. 21.37 of the first 1.32T tokens).
\end{inparaenum}

However, there are interesting exceptions:
\begin{inparaenum}[1)]
    \item Performance on prepositional phrase attachment (\texttt{PPA}) remains low, which is close to the random baseline, across all three stages, indicating that merely increasing the number of training tokens does not nessarily improve performance on this knowledge point.
    Even when examining a larger model, \texttt{Baichuan2\,13B}, we observe no significant performance gain on \texttt{PPA}.
    Nevertheless, fine-tuning for chat has been shown to improve performance on this particular knowledge point, as discussed in Finding~\ref{finding:chat_ppa}.
    The efficacy of other model families in learning \texttt{PPA} and the reasons why fine-tuning for chat is beneficial are intriguing topics for future research.
    \item The zero-shot performance on the knowledge point of indirect objects (\texttt{IO}) is substantially higher than few-shot performance from the 440B tokens' training stage onward.
    A closer investigation reveals that the model is confused and misled by in-context exemplars, tending to answer based on previous exemplars that it mistakenly associates with direct objects, which is more common than indirect objects.
    This tendency \textbf{to overvalue in-context exemplars at the expense of the question itself} is a phenomenon also observed in other smaller models, such as \texttt{Falcon\,1B/7B}, \texttt{Llama\,7B/13B}, and \texttt{Llama2\,7B/13B}, suggesting that smaller models may overly rely on in-context exemplars.
\end{inparaenum}

\section{Related Work}
\subsection{Evaluation of Large Language Models}
Recently, there has been a growing fascination with LLMs due to their remarkable performance across a wide spectrum of tasks.
Evaluating these models serves a dual purpose by revealing both their capabilities and limitations.
The results of the evaluation can offer valuable insights for refining and advancing LLMs.
Typically, evaluations are designed to assess the ability to perform specific tasks.
For example, GSM8k \citep{cobbe-etal-2021-training} evaluate the ability to perform mathematical reasoning, ToolLLM \citep{qin-etal-2023-toollm} evaluate the tool-use capabilities, and AGIEval \citep{zhong-etal-2023-agieval} use human-centric exams to evaluate the cognition and problem-solving abilities.

Besides task-specific evaluations, numerous evaluation benchmarks \citep{hendrycks-etal-2021-measuring,srivastava-etal-2022-beyond,liang-etal-2022-holistic} have been proposed to assess generalization capabilities of LLMs.
For example, HELM \citep{liang-etal-2022-holistic} evaluate prominent LLMs, covering a wide range of metrics, including model bias, efficiency, robustness, and more.

Our work belongs to the former category, specifically focusing on evaluating LLMs' linguistic comprehension capabilities.

\subsection{Syntactic Knowledge in Language Models}
Syntactic knowledge is a vast and complex topic, encompassing a wide range of aspects.
These include forming grammatically correct sentences, explaining specific syntactic phenomena, and deciphering the meaning of sentences.

Many previous studies have focused on the first two aspects.
They evaluate the syntactic knowledge of LLMs by constructing pairs of sentences, in which one syntactically acceptable and the other not.
The model's task is to determine which sentence is grammatically correct.
A representative work in this category is BLiMP \cite{warstadt-etal-2020-blimp}, covering 67 syntactic phenomena, including subject-verb agreement, filler-gap dependencies, and more.

In this work, we concentrate on the latter aspect: the ability to correctly understand the meaning of sentences.
There are previous studies in this direction that propose various methods, broadly categorized into \textbf{probing} and \textbf{prompting methods}.

\paragraph{Probing methods}
Probing methods are based on the premise that the syntactic knowledge required to understand a sentence should be reflected in the model's hidden states.
These methods aim to uncover and extract the latent hierarchical structure from a model's hidden layers, believed to represent syntactic knowledge \cite{hall-maudslay-etal-2020-tale, li-etal-2020-branching, newman-etal-2021-refining, zhao-etal-2023-transformers, kim-etal-2023-reconstruction}.
A probe is essentially a function, such as a static similarity metric or a trainable neural network, that measures the syntactic distance between two tokens.
If this distance is small, the token pair is considered to have a syntactic relationship or belong to the same constituent.
However, probing methods are limited to models with accessible hidden states, making models like the GPT series, which are API-based, unsuitable for probing.

\paragraph{Prompting methods}
Prompting methods are more flexible and applicable to any model supporting text generation.
Most work in this category involves prompting the model to parse a sentence into a hierarchical structure containing the syntactic knowledge needed to understand the sentence \cite{roy-etal-2022-benchclamp,bai-etal-2023-constituency,lin-etal-2023-chatgpt}.
Designing effective prompts for complex syntactic tasks remains a challenge, often requiring constrained decoding methods to ensure the model's output is in the desired format \cite{roy-etal-2022-benchclamp}.
In contrast, our work employs a specific type of prompting: the natural language Q\&A paradigm, a recently mainstream and LLM-friendly evaluation method \cite{cobbe-etal-2021-training,hendrycks-etal-2021-measuring,zhong-etal-2023-agieval,huang-etal-2023-c-eval}.
Thus, we circumvent the need for designing complex prompts or decoding methods.
\section{Conclusions}
\label{sec:conclusions}
In this work, we propose investigating the syntactic knowledge of LLMs by asking them natural language question answering, aiming to answer the question of whether LLMs truly understand language or just mimic comprehension via pattern recognition and memorization.
We crafted a series of questions focusing on nine syntactic knowledge points  that are fundamental to sentence comprehension.
Our experiments across 24 models suggest that LLMs have a basic ability to understand syntax, but their ability to correctly answer questions is limited and inconsistent, with the notable exception of the state-of-the-art \texttt{GPT4}, which exhibits exceptional performance.
Additionally, we find that the performance of LLMs varies greatly across different syntactic knowledge points, with prepositional phrase attachment being the most difficult and adjectival modifier and indirect object the easiest.
Finally, we conduct a case study on \texttt{Baichuan2} to investigate the training dynamics of syntactic knowledge.
We observe that the majority of syntactic knowledge is learned during the early stages of training of training.
This observation suggests that simply increasing the training tokens may not be the `\textit{silver bullet}' for improving the comprehension ability of LLMs.
\section*{Limitations}
This study is subject to several limitations.

The primary limitation stems from the indirect nature of our methodology, which lacks direct access to the model's hidden states and attention mechanisms.
As such, it lacks the capability to inspect the model's `\textit{neurons}' to determine how syntactic knowledge is stored and represented.
However, this limitation is not unique to our work and is shared by the majority of existing studies on LLMs evaluation.

Moreover, the scope of our syntactic evaluation is confined to the English language, meaning that the findings may not be generalizable across different languages, such as Chinese.

Additionally, our investigation covers only a select set of nine syntactic knowledge points.
The field of syntax is vast, and numerous other phenomena warrant further examination to gain a comprehensive understanding of LLMs' capabilities.

Lastly, our experimental setup was limited to models with fewer than 70 billion parameters due to resource constraints.
Thus, the behaviors and performance of larger, potentially more capable models remain unexplored in our study.

\ifarxiv
    \section*{Acknowledgments}
Special thanks are due to Chen Gong for generously providing access to invaluable computing resources, which significantly contributed to this study.
\fi

\bibliography{custom}
\clearpage

\appendix
\appendix
\section{Original Question Distribution}
\label{sec:appendix:original_question_dist}
\begin{table*}[t!]
    \setlength{\tabcolsep}{8pt}
    \centering
    \scalebox{0.9}{%
        \begin{tabular}{lc|ccccc}
            \toprule
            \textbf{Syntactic Knowledge Points}                          & \textbf{Abbr.}          & \textbf{\#TF}                  & \textbf{\#MC}     & \textbf{\#FITB}   & \textbf{\#total}      & \textbf{Ratio}\,(\texttt{\%})       \\
            \midrule
            \textbf{G}rammatical \textbf{S}ubject                        & \textbf{\texttt{GS}}    & 426,832                        & 106,708           & 106,708           & \phantom{0,}640,248   & \cellcolor{table_blue!44.79}14.93   \\
            \textbf{S}ubject \textbf{C}omplement                         & \textbf{\texttt{SC}}    & \wz59,984                      & \wz14,996         & \wz14,996         & \phantom{00,}89,976   & \cellcolor{table_blue!6.3}\wz2.10   \\
            \textbf{D}irect \textbf{O}bject                              & \textbf{\texttt{DO}}    & 261,320                        & \wz65,330         & \wz65,330         & \phantom{0,}391,980   & \cellcolor{table_blue!27.42}\wz9.14 \\
            \textbf{I}ndirect \textbf{O}bject                            & \textbf{\texttt{IO}}    & \wz\wz2,716                    & \phantom{000,}679 & \phantom{000,}679 & \phantom{000,}4,074   & \cellcolor{table_blue!0.27}\wz0.09  \\
            \textbf{M}ain \textbf{V}erb \textbf{P}hrase                  & \textbf{\texttt{MVP}}   & 750,852$^\ddagger$\kern-0.45em & 129,669           & 129,669           & 1,010,190             & \cellcolor{table_blue!70.65}23.55   \\
            \textbf{ADJ}ectival modifier$^\dagger$                       & \textbf{\texttt{ADJ}}   & 587,968                        & \wz67,865         & \wz58,401         & \phantom{0,}714,234   & \cellcolor{table_blue!49.95}16.65   \\
            \textbf{ADV}erbial modifier (Adjunct)                        & \textbf{\texttt{ADV}}   & 385,406                        & \wz77,439         & \wz40,268         & \phantom{0,}503,113   & \cellcolor{table_blue!35.19}11.73   \\
            \midrule
            \textbf{CO}ordination                                        & \textbf{\texttt{CO}}    & 319,492                        & \wz33,405         & \wz19,594         & \phantom{0,}372,491   & \cellcolor{table_blue!26.04}\wz8.68 \\
            {\textbf{P}repositional \textbf{P}hrase \textbf{A}ttachment} & {\textbf{\texttt{PPA}}} & {375,576}                      & {\wz93,894}       & {\wz93,894}       & {\phantom{0,}563,364} & \cellcolor{table_blue!39.39}{13.13} \\
            \bottomrule
        \end{tabular}
    }
    \caption{Syntactic knowledge points in our evaluation. $^\dagger$: We only consider post-modifier, such as relative clause and reduced relative clause in this work. $^\ddagger$: The questions of main verb phrase in True/False are the same as those in surface subject, subject complement, direct object, and indirect object, so we directly reuse the questions of these four syntactic knowledge points and do not count them in the total number of questions.}
    \label{tab:syntactic_knowledge_dist}
\end{table*}
The original distribution of the questions we built is shown in Table~\ref{tab:syntactic_knowledge_dist}.
From this table, we can see that the distribution of each syntactic knowledge point is imbalanced.
The most common syntactic knowledge point is the main verb phrase, which accounts for 23.55\% of all the questions, while the least common syntactic knowledge point is the indirect object, which only accounts for 0.09\%.

\begin{table*}[tp!]
    \centering
    \scalebox{0.85}{%
        \begin{tabular}{p{8cm}ccc}
            \toprule
            \textbf{Model}                          & \textbf{Creator}            & \textbf{\#Parameters}               & \textbf{Open-sourced} \\
            \midrule
            \rowcolor[gray]{.95}
            Mistral series                          &                             &                                     &                       \\
            \quad \texttt{Mistral-7B-v0.1}          & \multirow{2}{*}{Mistral AI} & \multirow{2}{*}{\texttt{\wz 7.24B}} & \multirow{2}{*}{\yes} \\
            \quad \texttt{Mistral-7B-Instruct-v0.1} &                             &                                     &                       \\
            \hline
            \rowcolor[gray]{.95}
            Baichuan2 series                        &                             &                                     &                       \\
            \quad \texttt{Baichuan2-7B-Base}        & \multirow{4}{*}{Baichuan}   & \multirow{2}{*}{\texttt{\wz 7.51B}} & \multirow{4}{*}{\yes} \\
            \quad \texttt{Baichuan2-7B-Chat}        &                             &                                     &                       \\
            \quad \texttt{Baichuan2-13B-Base}       &                             & \multirow{2}{*}{\texttt{13.90B}}    &                       \\
            \quad \texttt{Baichuan2-13B-Chat}       &                             &                                     &                       \\
            \hline
            \rowcolor[gray]{.95}
            Falcon series                           &                             &                                     &                       \\
            \quad \texttt{falcon-rw-1b}             & \multirow{5}{*}{TII}        & \texttt{\wz 1.31B}                  & \multirow{5}{*}{\yes} \\
            \quad \texttt{falcon-7b}                &                             & \multirow{2}{*}{\texttt{\wz 6.92B}} &                       \\
            \quad \texttt{falcon-7b-instruct}       &                             &                                     &                       \\
            \quad \texttt{falcon-40b}               &                             & \multirow{2}{*}{\texttt{41.30B}}    &                       \\
            \quad \texttt{falcon-40b-instruct}      &                             &                                     &                       \\
            \hline
            \rowcolor[gray]{.95}
            LLaMA series                            &                             &                                     &                       \\
            \quad \texttt{llama-7b}                 & \multirow{4}{*}{Meta}       & \texttt{\wz 6.78B}                  & \multirow{4}{*}{\yes} \\
            \quad \texttt{llama-13b}                &                             & \texttt{13.02B}                     &                       \\
            \quad \texttt{llama-30b}                &                             & \texttt{32.53B}                     &                       \\
            \quad \texttt{llama-65b}                &                             & \texttt{65.29B}                     &                       \\
            \hline
            \rowcolor[gray]{.95}
            LLaMA2 series                           &                             &                                     &                       \\
            \quad \texttt{llama-2-7b}               & \multirow{6}{*}{Meta}       & \multirow{2}{*}{\texttt{\wz 6.74B}} & \multirow{6}{*}{\yes} \\
            \quad \texttt{llama-2-7b-chat}          &                             &                                     &                       \\
            \quad \texttt{llama-2-13b}              &                             & \multirow{2}{*}{\texttt{13.02B}}    &                       \\
            \quad \texttt{llama-2-13b-chat}         &                             &                                     &                       \\
            \quad \texttt{llama-2-70b}              &                             & \multirow{2}{*}{\texttt{68.98B}}    &                       \\
            \quad \texttt{llama-2-70b-chat}         &                             &                                     &                       \\
            \hline
            \rowcolor[gray]{.95}
            ChatGPT series                          &                             &                                     &                       \\
            \quad \texttt{gpt-3.5-turbo-0613}       & \multirow{2}{*}{OpenAI}     & \multirow{2}{*}{\texttt{unknown}}   & \multirow{2}{*}{\no}  \\
            \quad \texttt{gpt-4-0613}               &                             &                                     &                       \\
            \bottomrule
        \end{tabular}
    }
    \caption{
        Models evaluated in this work.
        ``\#Parameters'' is the number of parameters of the model.
        ``Open-sourced'' indicates whether the model is open sourced.
    }
    \label{tab:models}
\end{table*}
\section{Model Details}
\label{sec:appendix:model_details}
The information about the models we evaluated in this work is shown in Table~\ref{tab:models}.

\paragraph{Mistral series: }
Mistral is Mistral AI’s first Large Language Model (LLM), a transformer model especially suited for NLP applications.
It’s trained on a vast dataset of text and code, enabling it to generate text, translate languages, produce creative content, and answer questions instructively.
Mistral 7B, with 7.24 billion parameters, outperforms LLaMA 2 13B on all benchmarks and LLaMA 30B on many other benchmarks.

\paragraph{Baichuan2 series: }
The newest open-source and commercially available large language model series from Baichuan Inc.
This series comprises four models: a 7B and a 13B foundation model, each with their corresponding chat versions.
The \texttt{Baichuan2\,7B} model is one of the few models that publicly release intermediate checkpoints, which facilitates our case study of the training dynamics of syntactic knowledge.

\paragraph{Falcon series: }
A series of large language models published by TII, trained on the Refined Web Dataset.
This series includes three models with parameter sizes of 1B, 7B, and 40B.
The 7B and 40B versions also have their corresponding instruction-tuned variants.

\paragraph{LLaMA series: }
One of the most popular large language model series from Meta, which has been used in various works.
This series includes four models with parameter sizes of 7B, 13B, 30B, and 65B.

\paragraph{LLaMA2 series: }
The new generation of the LLaMA series, trained on a cleaner and larger dataset.
This series consists of three models with parameter sizes of 7B, 13B, and 70B, each with their corresponding chat versions.

\paragraph{ChatGPT series: }
Currently regarded as the most powerful large language model series, developed by OpenAI.
However, most models in this series are accessible as pay-to-use, API-only models. For our experiments, we focused on two chat versions from this series: `\texttt{gpt-3.5-turbo-0613}' and `\texttt{gpt-4-0613}'.

\subsection{Implementation Details}
For GPT series, we use the official Python API to access the models.
We set the temperature to 0 and maximum length to 256 for ``Fill in the Blank'' questions and 10 for ``True/False'' and ``Multiple Choice'' questions.
Other hyper-parameters are remained as default.

For other open-sourced models, we use the \texttt{transformers} library~\cite{wolf-etal-2020-transformers} to access them.
We \textbf{do not fine-tune} any of these models.
If the model creator provides the special generation function, such as ``\texttt{chat()}'' in the \texttt{Baichuan2} series, we directly use it, otherwise we use the ``\texttt{generate()}'' function.
The hyper-parameters are set to the same as the GPT series.

We use the same prompt for all the models, if the model creator does not provide a suggested prompt.

In few-shot experiments, for each question, we randomly select 5 exemplars having the same syntactic knowledge point and question type as the question has.

We run all the experiments with three random seeds, which will affect the exemplars selected for each question, and report the average results.
The only exception is that we only run with one random seed on the pay-to-use GPT models, due to the high price of using them.

\subsection{The Problem of CoT}
\label{sec:appendix:cot}
Our decision to exclude the CoT setting is grounded in two primary reasons.
Firstly, in most instances, discerning the syntactic structure of a sentence does not require complex reasoning.
Secondly, preliminary tests revealed that many models, particularly the less complex ones, struggled to generate coherent chains of thought tailored to our syntactic knowledge questions.
Often, these models repetitively produce phrases like "The object of the \texttt{XXX} is \texttt{YYY}," extending up to the preset maximum generation length.

\section{Detailed Results}
\label{sec:appendix:detailed_results}
The results of all the models under all the settings are shown in Table~\ref{tab:main_results_all}.
The correlation between the difficulty metrics is shown in Table~\ref{tab:correlation}.
The overall accuracy of each model under each zero-shot and few-shot setting is shown in Table~\ref{tab:acc_vs_kp_oa_all_zero-shot} and Table~\ref{tab:acc_vs_kp_all_oa}, respectively.
The detailed performance of each model under each zero-shot and few-shot setting is shown in Table~\ref{tab:acc_vs_kp_all_zero-shot} and Table~\ref{tab:acc_vs_kp_all}, respectively.

\begin{table*}[tb]
    \centering
    \scalebox{0.85}{%
\begin{tabular}{lccccccccccc}
            \toprule
                                                                       & \multicolumn{5}{c}{\textbf{Zero-shot}}      &                                             & \multicolumn{5}{c}{\textbf{Few-shot}}                                                                                                                                                                                                                                                                                                                                                       \\
            \cmidrule(l{0pt}r{0pt}){2-6}\cmidrule(l{0pt}r{0pt}){8-12}  & \textbf{TF}                                 & \textbf{MC}                                 & \multicolumn{2}{c}{\textbf{FITB}}           & \multirow{2}{*}{\textbf{OA}}                &                                             & \textbf{TF} & \textbf{MC}                                 & \multicolumn{2}{c}{\textbf{FITB}}           & \multirow{2}{*}{\textbf{OA}}                                                                                                            \\
            \cmidrule(l{0pt}r{0pt}){4-5}\cmidrule(l{0pt}r{0pt}){10-11} & Acc.                                        & Acc.                                        & Acc.                                        & $F_1$                                       &                                             &             & Acc.                                        & Acc.                                        & Acc.                                        & $F_1$                                       &                                             \\
\bottomrule
        \end{tabular}
    }
    \caption{Main results of our evaluation.}
    \label{tab:main_results_all}
\end{table*}

\begin{table*}[tb]
    \centering
    \scalebox{0.85}{%
        \begin{tabular}{llcccccccccc}
            \toprule
                                     &                                &       & \multicolumn{4}{c}{\textbf{Zero-shot}} &                                     & \multicolumn{4}{c}{\textbf{Few-shot}}                                                                                                                                                                                                             \\
            \cmidrule(l{0pt}r{0pt}){4-7}\cmidrule(l{0pt}r{0pt}){9-12}
                                     &                                &       & \textbf{TF}                            & \textbf{MC}                         & \multicolumn{2}{c}{\textbf{FITB}}     &                                     & \textbf{TF} & \textbf{MC}                         & \multicolumn{2}{c}{\textbf{FITB}}                                                                               \\
            \cmidrule(l{0pt}r{0pt}){6-7}\cmidrule(l{0pt}r{0pt}){11-12}
                                     &                                &       & Acc.                                   & Acc.                                & Acc.                                  & $F_1$                               &             & Acc.                                & Acc.                                & Acc.                                & $F_1$                               \\
            \midrule
            \multirow{4}{*}{Kendall} & \textbf{TF}                    & Acc.  & \cellcolor{gray!20}                    & \cellcolor{table_blue!50.400} 0.560 & \cellcolor{table_blue!43.200} 0.480   & \cellcolor{table_blue!46.473} 0.516 &             & \cellcolor{gray!20}                 & \cellcolor{table_blue!60.652} 0.674 & \cellcolor{table_blue!44.102} 0.490 & \cellcolor{table_blue!42.391} 0.471 \\
\bottomrule
        \end{tabular}
    }
    \caption{The correlation coefficient between the metrics.}
    \label{tab:correlation}
\end{table*}

\begin{table*}[tb]
    \setlength{\tabcolsep}{6pt}
    \centering
    \scalebox{0.85}{%
        \begin{tabular}{lccccccccc}
            \toprule
            \textbf{Models}                 & \textbf{GS}                                 & \textbf{SC}                                 & \textbf{DO}                                 & \textbf{IO}                                 & \textbf{MVP}                                & \textbf{ADJ}                                & \textbf{ADV}                                & \textbf{PPA}                                & \textbf{CO}                                 \\
            \midrule
            \texttt{Mistral 7B}             & \cellcolor{table_blue!28.382}58.75          & \cellcolor{table_blue!24.910}53.49          & \cellcolor{table_blue!23.469}54.71          & \cellcolor{table_blue!30.650}60.79          & \cellcolor{table_blue!21.143}50.11          & \cellcolor{table_blue!16.960}47.73          & \cellcolor{table_blue!15.120}43.97          & \cellcolor{table_blue!5.028}30.87           & \cellcolor{table_blue!27.538}56.69          \\
            \midrule
            \texttt{Baichuan2 7B}           & \cellcolor{table_blue!23.713}53.98          & \cellcolor{table_blue!16.807}44.82          & \cellcolor{table_blue!18.295}49.56          & \cellcolor{table_blue!38.207}68.57          & \cellcolor{table_blue!23.406}52.49          & \cellcolor{table_blue!15.469}46.23          & \cellcolor{table_blue!14.930}43.77          & \cellcolor{table_blue!1.331}26.83           & \cellcolor{table_blue!20.550}49.31          \\
            \texttt{Baichuan2 13B}          & \cellcolor{table_blue!27.276}57.62          & \cellcolor{table_blue!25.379}53.99          & \cellcolor{table_blue!26.529}57.75          & \cellcolor{table_blue!32.825}63.03          & \cellcolor{table_blue!24.958}54.13          & \cellcolor{table_blue!15.147}45.90          & \cellcolor{table_blue!18.502}47.52          & \cellcolor{table_blue!1.417}26.93           & \cellcolor{table_blue!27.544}56.70          \\
\bottomrule
        \end{tabular}
    }
    \caption{Overall performance of each model under \textbf{Zero}-shot setting at the knowledge point level.}
    \label{tab:acc_vs_kp_oa_all_zero-shot}
\end{table*}

\begin{table*}[tb]
    \setlength{\tabcolsep}{6pt}
    \centering
    \scalebox{0.85}{%
        \begin{tabular}{lccccccccc}
            \toprule
            \textbf{Models}                 & \textbf{GS}                                 & \textbf{SC}                                 & \textbf{DO}                                 & \textbf{IO}                                 & \textbf{MVP}                                & \textbf{ADJ}                                & \textbf{ADV}                                & \textbf{PPA}                                & \textbf{CO}                                 \\
            \midrule
            \texttt{Mistral 7B}             & \cellcolor{table_blue!33.048}62.81          & \cellcolor{table_blue!27.828}57.68          & \cellcolor{table_blue!32.422}63.22          & \cellcolor{table_blue!38.099}68.76          & \cellcolor{table_blue!30.500}59.66          & \cellcolor{table_blue!34.070}64.06          & \cellcolor{table_blue!25.116}55.13          & \cellcolor{table_blue!9.480}38.26           & \cellcolor{table_blue!31.108}60.74          \\
            \midrule
            \texttt{Baichuan2 7B}           & \cellcolor{table_blue!27.134}56.62          & \cellcolor{table_blue!22.599}52.26          & \cellcolor{table_blue!24.266}55.01          & \cellcolor{table_blue!25.700}56.16          & \cellcolor{table_blue!25.012}53.83          & \cellcolor{table_blue!30.087}59.92          & \cellcolor{table_blue!21.297}51.21          & \cellcolor{table_blue!2.312}30.78           & \cellcolor{table_blue!26.537}55.95          \\
            \texttt{Baichuan2 13B}          & \cellcolor{table_blue!29.989}59.61          & \cellcolor{table_blue!28.780}58.66          & \cellcolor{table_blue!30.622}61.41          & \cellcolor{table_blue!37.258}67.90          & \cellcolor{table_blue!31.461}60.68          & \cellcolor{table_blue!32.234}62.15          & \cellcolor{table_blue!24.393}54.39          & \cellcolor{table_blue!1.994}30.45           & \cellcolor{table_blue!26.548}55.96          \\
\bottomrule
        \end{tabular}
    }
    \caption{Overall performance of each model under Few-shot setting at the knowledge point level.}
    \label{tab:acc_vs_kp_all_oa}
\end{table*}

\begin{table*}[tb]
    \setlength{\tabcolsep}{6pt}
    \centering
    \scalebox{0.85}{%
        \begin{tabular}{llccccccccc}
            \toprule
            \textbf{Q. Types}                    & \textbf{Models}                 & \textbf{GS}                                 & \textbf{SC}                                 & \textbf{DO}                                 & \textbf{IO}                                 & \textbf{MVP}                                & \textbf{ADJ}                                & \textbf{ADV}                                & \textbf{PPA}                                & \textbf{CO}                                 \\
            \midrule
            \multirow{23}{*}{\textbf{TF} (Acc.)} & \texttt{Mistral 7B}             & \cellcolor{table_blue!5.783}56.15           & \cellcolor{table_blue!16.026}56.15          & \cellcolor{table_red!2.629}53.33            & \cellcolor{table_red!15.918}40.00           & \cellcolor{table_blue!3.982}54.09           & \cellcolor{table_red!1.484}49.73            & \cellcolor{table_red!9.013}40.00            & \cellcolor{table_blue!4.898}50.91           & \cellcolor{table_blue!11.413}55.76          \\
            \cmidrule(l{0pt}r{0pt}){2-11}
                                                 & \texttt{Baichuan2 7B}           & \cellcolor{table_blue!7.229}56.92           & \cellcolor{table_blue!0.782}46.15           & \cellcolor{table_blue!3.194}57.33           & \cellcolor{table_blue!30.545}70.00          & \cellcolor{table_blue!5.217}54.77           & \cellcolor{table_red!1.484}49.73            & \cellcolor{table_blue!15.659}56.97          & \cellcolor{table_blue!9.306}53.64           & \cellcolor{table_blue!8.392}53.94           \\
\end{tabular}
    }
    \caption{Performance of each model under \textbf{Zero}-shot setting at the knowledge point level.}
    \label{tab:acc_vs_kp_all_zero-shot}
\end{table*}
\begin{table*}[tb]
    \setlength{\tabcolsep}{6pt}
    \centering
    \scalebox{0.85}{%
        \begin{tabular}{llccccccccc}
            \rowcolor[gray]{.9}
            \multicolumn{11}{c}{continued from previous page}                                                                                                                                                                                                                                                                                                                                                                                                                                                       \\
            \midrule
            \textbf{Q. Types}                       & \textbf{Models}                 & \textbf{GS}                                 & \textbf{SC}                                 & \textbf{DO}                                 & \textbf{IO}                                 & \textbf{MVP}                                & \textbf{ADJ}                                & \textbf{ADV}                                & \textbf{PPA}                                & \textbf{CO}                                 \\
            \midrule
            \multirow{23}{*}{\textbf{FITB} (Acc.)}  & \texttt{Mistral 7B}             & \cellcolor{table_blue!37.049}56.19          & \cellcolor{table_blue!32.461}49.41          & \cellcolor{table_blue!26.949}41.38          & \cellcolor{table_blue!49.451}75.00          & \cellcolor{table_blue!24.047}37.06          & \cellcolor{table_blue!18.446}28.15          & \cellcolor{table_blue!17.774}26.96          & \cellcolor{table_blue!8.382}13.00           & \cellcolor{table_blue!39.560}60.00          \\
\bottomrule
        \end{tabular}
    }
    \caption{Performance of each model under \textbf{Zero}-shot setting at the knowledge point level (Continued).}
\end{table*}

\clearpage
\begin{table*}[tb]
    \setlength{\tabcolsep}{6pt}
    \centering
    \scalebox{0.85}{%
        \begin{tabular}{llccccccccc}
            \toprule
            \textbf{Q. Types} & \textbf{Models}                 & \textbf{GS}                                 & \textbf{SC}                                 & \textbf{DO}                                 & \textbf{IO}                                  & \textbf{MVP}                                & \textbf{ADJ}                                & \textbf{ADV}                                & \textbf{PPA}                                & \textbf{CO}                                 \\
            \midrule
            \multirow{23}{*}{\textbf{TF} (Acc.)}
                              & \texttt{Mistral 7B}             & \cellcolor{table_blue!12.042}55.38          & \cellcolor{table_blue!8.057}53.59           & \cellcolor{table_blue!12.733}57.78          & \cellcolor{table_blue!5.538}52.22            & \cellcolor{table_blue!10.633}55.45          & \cellcolor{table_blue!7.663}54.23           & \cellcolor{table_blue!4.891}53.13           & \cellcolor{table_blue!9.954}57.58           & \cellcolor{table_blue!27.309}64.44          \\
            \cmidrule(l{0pt}r{0pt}){2-11}
                              & \texttt{Baichuan2 7B}           & \cellcolor{table_blue!7.889}52.82           & \cellcolor{table_blue!0.848}49.23           & \cellcolor{table_red!0.264}50.22            & \cellcolor{table_blue!7.385}53.33            & \cellcolor{table_blue!3.038}50.91           & \cellcolor{table_red!0.000}49.73            & \cellcolor{table_red!4.337}46.67            & \cellcolor{table_red!3.837}48.79            & \cellcolor{table_blue!13.815}55.96          \\
\end{tabular}
    }
    \caption{Performance of each model under Few-shot setting at the knowledge point level.}
    \label{tab:acc_vs_kp_all}
\end{table*}
\begin{table*}[tb]
    \setlength{\tabcolsep}{6pt}
    \centering
    \scalebox{0.85}{%
        \begin{tabular}{llccccccccc}
            \rowcolor[gray]{.9}
            \multicolumn{11}{c}{continued from previous page}                                                                                                                                                                                                                                                                                                                                                                                                                                  \\
            \midrule
            \textbf{Q. Types} & \textbf{Models}                 & \textbf{GS}                                 & \textbf{SC}                                 & \textbf{DO}                                 & \textbf{IO}                                  & \textbf{MVP}                                & \textbf{ADJ}                                & \textbf{ADV}                                & \textbf{PPA}                                & \textbf{CO}                                 \\
            \midrule
            \multirow{23}{*}{\textbf{FITB} (Acc.)}
                              & \texttt{Mistral 7B}             & \cellcolor{table_blue!40.892}62.22          & \cellcolor{table_blue!34.949}53.33          & \cellcolor{table_blue!32.297}49.20          & \cellcolor{table_blue!55.970}85.00           & \cellcolor{table_blue!34.840}53.33          & \cellcolor{table_blue!52.174}79.26          & \cellcolor{table_blue!26.947}40.87          & \cellcolor{table_blue!18.750}28.67          & \cellcolor{table_blue!42.639}64.73          \\
\bottomrule
        \end{tabular}
    }
    \caption{Performance of each model under Few-shot setting at the knowledge point level (Continued).}

\end{table*}

\input{figures/acc_vs_steps_zero-shot}

\end{document}